\documentclass[runningheads]{llncs}

\usepackage{eccv}

\usepackage{eccvabbrv}

\usepackage{graphicx}
\graphicspath{{./imgs/}}
\usepackage{booktabs}

\usepackage[accsupp]{axessibility}  

\usepackage{hyperref}

\usepackage{orcidlink}
\usepackage{multirow}
\usepackage{placeins}
\usepackage{array}
\usepackage{wrapfig}

\begin{document}

\title{ModaFlow: Modality-Aware Flow Matching for High-Fidelity Virtual Try-On} 

\titlerunning{ModaFlow}

\author{Xiangyu Sai\inst{1}\orcidlink{0009-0002-4407-7328} \and
Meysam Madadi\inst{2,3}\orcidlink{0000-0002-7384-5712} \and
Sergio Escalera\inst{2,3}\orcidlink{0000-0003-0617-8873}\and
Yong Xu\inst{1,4}\orcidlink{0000-0001-7183-3155}}

\authorrunning{X.~Sai et al.}

\institute{South China University of Technology, Guangzhou, China \and
Universitat de Barcelona, Barcelona, Spain \and
Computer Vision Center, Bellaterra, Spain \and
Guangdong Provincial Key Laboratory of Multimodal Big Data Intelligent Analysis, Guangzhou, China}

\maketitle

\begin{center}
\small arXiv preprint, 2026
\end{center}

\FloatBarrier

\begin{figure}[]
\centering
   \includegraphics[width=0.9\textwidth]{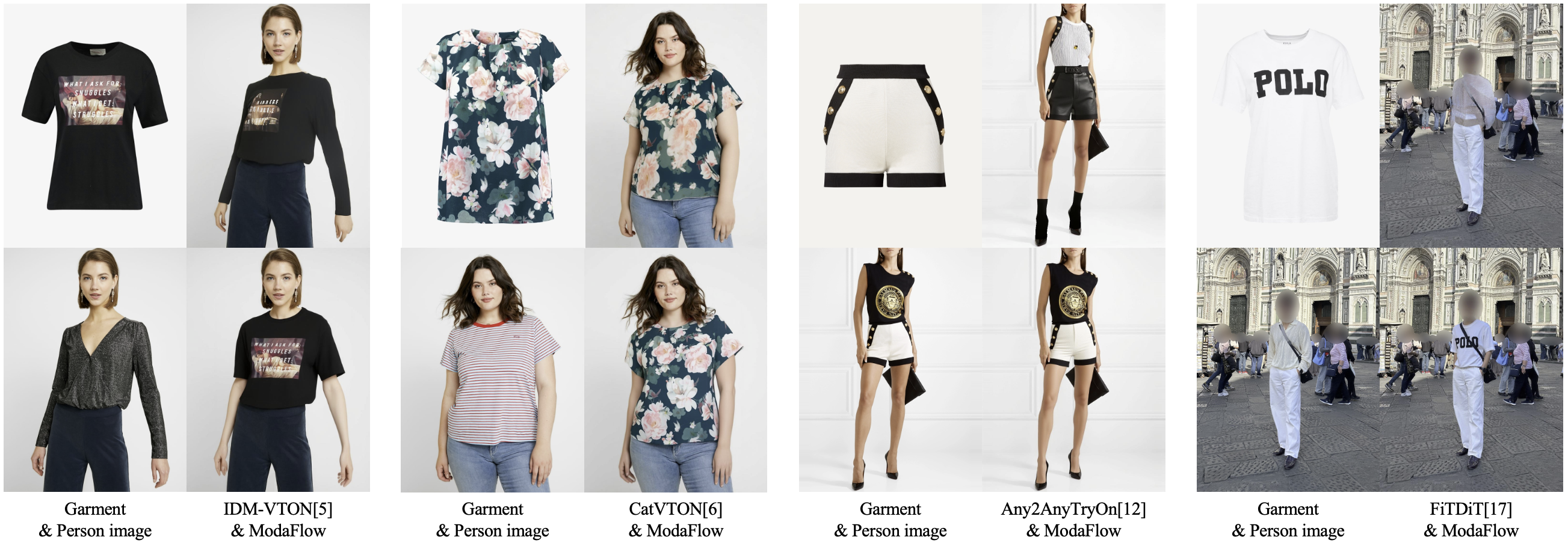}
   \caption{Qualitative comparisons on four challenging cases \textbf{left to right}: \textbf{(1)} text/logo, \textbf{(2)} patterns, \textbf{(3)} bottoms, \textbf{(4)} in-the-wild.
   \textbf{Top}: representative recent SOTAs show visible artifacts or misalignment; \textbf{Bottom}: \textbf{ModaFlow} produces more plausible results with better detail preservation.}
   \label{fig:fig1}
\end{figure}
\vspace{-30pt}
\begin{abstract}
Image-based virtual try-on has emerged as a compelling task in e-commerce and augmented reality, yet existing methods struggle to simultaneously preserve fine garment semantics and adapt to diverse person body geometries under large clothing–body deformations.
We present ModaFlow, a modality-aware flow-matching based framework for high-fidelity virtual try-on that achieves precise alignment between textual descriptions and garment appearance.
Unlike prior methods that treat multimodal conditions uniformly, ModaFlow introduces a modality-aware guidance scheme: visual garment embeddings extracted by a pretrained image prompt adapter provide deterministic, persistent structural guidance, while textual embeddings generated from garment descriptions are controlled via classifier-free guidance (CFG) with adaptive scaling and zero-initialized velocity.
To further enhance flow field accuracy, we propose two regularization losses—cosine similarity and perceptual flow discrimination—that jointly improve directional consistency and perceptual realism of the velocity field.
Additionally, a mask manipulation strategy stochastically samples among box, transparent, and relaxed masks during training, simulating diverse occlusion scenarios and enabling robust inference under unpaired settings where only a box mask is available.
Experiments show that ModaFlow achieves state-of-the-art results in both qualitative and quantitative evaluations, reducing FID by approximately 30\% on paired and 20\% on unpaired benchmarks.
\keywords{Virtual Try-On \and Flow Matching \and Image Synthesis}
\end{abstract} 
\section{Introduction}
\label{sec:intro}

Image-based virtual try-on aims to synthesize a realistic image of a person wearing a target garment, and has become an essential component in e-commerce, digital fashion, and augmented reality. 
A successful try-on system should faithfully preserve garment semantics (\eg, texture, pattern, and style) while maintaining the person’s pose, body shape, and identity. 
However, these objectives are often in conflict: garments exhibit complex deformations across diverse human geometries, and visual cues can be partially occluded or misaligned in unpaired real-world settings.

Early approaches based on generative adversarial networks (GANs)~\cite{viton,cpvton,acgpn,tryongan} demonstrated the feasibility of virtual try-on but often produce artifacts and texture loss due to limited spatial reasoning and unstable training. 
Recent diffusion-based models~\cite{stableviton,idmvton,catvton,mcvton,boow,fitdit,tpd,ootd} improve visual realism and garment fidelity, yet remain treating visual and textual conditions uniformly, leading to misalignment between description and appearance.
Moreover, most diffusion-based methods rely on paired training data and fixed deformation fields, limiting their adaptability to unpaired scenarios with diverse body geometries.

To address these challenges, we propose \textit{ModaFlow}, a flow-matching based framework for high-fidelity virtual try-on that unifies visual and textual guidance. 
We introduce a \textit{modality-aware guidance} mechanism that asymmetrically integrates two condition streams: 
visual garment embeddings from image prompt adapter, which provide deterministic, persistence structural guidance, 
and controllable textual embeddings from text encoder model, which deliver classifier-free guidance with adaptive scaling and zero-initialized velocity, enabling precise semantic alignment.
To ensure stable and perceptually consistent flow learning, we further introduce two regularization objectives, cosine similarity and perceptual flow discrimination, that jointly enhance geometric alignment and flow realism. 
Finally, our \textit{mask manipulation strategy} with stochastically employing box, transparent, and relaxed masks during training exposes the model to diverse occlusion conditions, allowing it to generalize across varying body geometries and unpaired try-on settings.
As shown in Figure~\ref{fig:fig1}, ModaFlow produces high-fidelity try-on results with accurate garment–pose alignment and fine-grained texture preservation.

The key contributions of this work are:
\begin{enumerate}
    \item A flow-matching framework for virtual try-on that jointly optimizes garment semantics and body geometry through triptych reconstruction, enabling high-fidelity synthesis under unpaired settings.
    \item A modality-aware guidance design combining deterministic visual conditioning with adaptive textual guidance, enabling fine-grained text-garment alignment while balancing structural precision and semantic expressivity and controllability.
    \item A novel mask manipulation strategy, specifically designed for cross-garment category failures for unpaired settings common in state-of-the-art, allowing the input garment to adapt naturally to diverse scenarios.
    \item Two flow regularization losses, cosine similarity and perceptual flow discrimination, that jointly improve the directional consistency and perceptual realism of the learned velocity field.
    \item State-of-the-art performance on multiple benchmarks, with substantial improvements in garment fidelity, text–garment alignment, and geometric adaptation.
\end{enumerate}


\section{Related work}
\label{sec:related}

We review the most relevant literature in terms of the evolution of \textit{image-based virtual try-on} methods, \textit{diffusion and flow-matching models} that form the backbone of recent generative approaches, and \textit{guidance strategies}, which are crucial for controllable and multimodal generation.

\begin{figure}[t]
\centering
   \fbox{\includegraphics[width=0.95\linewidth]{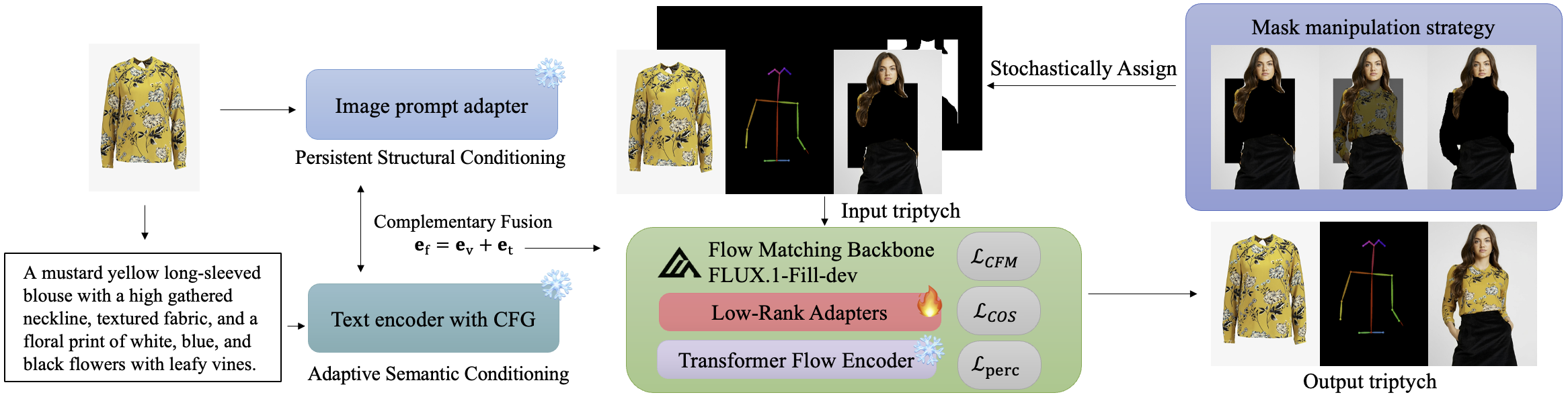}}
   \caption{ModaFlow pipeline. The model takes a triptych input consisting of a garment image, OpenPose map, and garment-agnostic person image generated through our stochastic mask manipulation strategy. Visual garment embeddings $\mathbf{e}_v$ from the image prompt adapter provide persistent structural conditioning, while textual semantics $\mathbf{e}_t$ are encoded with classifier-free guidance. The two modalities are fused through complementary fusion and injected into the flow-matching backbone, equipped with LoRA fine-tuning and two flow-regularization losses. The model outputs a triptych with aligned pose and high-fidelity garment details.}
   \label{fig:figpipe}
\end{figure}
\subsection{Image Based Virtual Try-on}

Image-based virtual try-on seeks to synthesize a photorealistic image of a person wearing a target garment, given separate images of the person and the clothing.
Early methods adopt a two-stage GAN-based pipeline: VITON~\cite{viton} first warps the garment and then refines the composition, while CP-VTON~\cite{cpvton} improves alignment using a learnable thin-plate spline transformation to better preserve geometric alignment.
CIT~\cite{cit} inserted Transformer blocks into the U-Net generator on top of the CP-VTON framework to enhance long-range context.
ACGPN~\cite{acgpn} further addresses occlusions by predicting a semantic layout and adaptively deciding whether to generate or preserve content per body region.
Instead of deforming the garment pixels, TryOnGAN~\cite{tryongan} maps both person and clothing into StyleGAN’s latent space and perform the try on by manipulating latent codes, yielding very coherent results within the StyleGAN’s learned manifold . 

The rise of diffusion models has enabled higher-fidelity synthesis.
StableVITON~\cite{stableviton} leverages a pre-trained latent diffusion model and learns semantic correspondence via zero cross-attention to retain garment details. 
IDM-VTON~\cite{idmvton} enhances identity preservation by fusing multi-level garment features into both cross- and self-attention layers of the denoising UNet.
OOTDiffusion~\cite{ootd} demonstrates the advantages of diffusion models in controllability. The garment’s features are injected directly into the diffusion model’s layers, and an adaptation of classifier-free guidance is used during training.
However, many such methods rely on auxiliary networks (\eg, ReferenceNet) or complex conditioning (pose, parsing, text), increasing computational overhead and input requirements.

Recent work emphasizes efficiency and robustness in unconstrained settings.
CatVTON~\cite{catvton} shows that simple spatial concatenation of inputs suffices, eliminating auxiliary networks and complex conditioning.
Texture-Preserving Diffusion (TPD)~\cite{tpd} introduces a warping-free diffusion-based framework that leverages the inherent self-attention layers within the denoising UNet.
To handle real-world scenarios where precise masks are unavailable, BooW-VTON~\cite{boow} introduces a mask-free training paradigm, while FitDiT~\cite{fitdit} improves detail fidelity through frequency-aware learning and adaptive masking for cross-category fitting.
MC-VTON~\cite{mcvton} achieves strong performance using only the native Diffusion Transformer~\cite{dit} backbone with minimal fine-tuning.
Any2AnyTryOn~\cite{any2anytryon} addresses the scarcity of paired data by constructing a large-scale try-on dataset and enables instruction-driven try-on without relying on masks or poses via adaptive position embeddings.
VTON-HandFit~\cite{handfit} focuses on hand-occlusion cases by incorporating explicit hand/pose priors and disentangled hand embeddings to recover occluded garment structure and appearance.
DualFit~\cite{dualfit} adopts a two-stage hybrid pipeline that first aligns garments via a learned flow field and then synthesizes the final result with preserved-region guidance to better retain high-frequency details such as logos and printed text.

ModaFlow adopts a flow-matching paradigm with modality-aware conditioning, eliminating warping and auxiliary encoders while achieving high fidelity and adaptability across diverse geometries.

\subsection{Diffusion and Flow Matching Models}

Generative Adversarial Networks (GANs)~\cite{gan,styleganxl} have long dominated image synthesis. However, their training instability and sensitivity to hyperparameters often limit scalability and sample diversity.

Denoising Diffusion Probabilistic Models (DDPM)~\cite{ddpm} introduced a new paradigm by gradually denoising data from a Gaussian prior, enabling stable training and high-fidelity generation.
Subsequent work demonstrated that diffusion models can surpass GANs in both quality and robustness~\cite{cg}, especially when scaled with powerful architectures.
For instance, DiT~\cite{dit} and SDXL~\cite{sdxl} leveraged Transformers and latent-space diffusion for high-resolution synthesis, while SD3~\cite{sd3} further scaled Rectified flow transformers for improved performance.
Beyond architectural scaling, recent efforts address artifacts and distortions with multi-resolution modeling and adaptive normalization~\cite{dimr}.
Also, Score Distillation Sampling was initially proposed in DreamFusion~\cite{dreamfusion} to tackle text-to-3D synthesis. They use a pretrained text-to-image diffusion model as a guidance “critic” to train another generative system.

More recently, \textit{flow matching}~\cite{fm,recflow,nf} has emerged as a compelling alternative that replaces the stochastic diffusion process with a deterministic ordinary differential equation (ODE).
Instead of modeling noise prediction, flow matching directly learns a continuous velocity field $v_\theta(\mathbf{x}, t)$ that transports samples from noise to data along straight or near-straight trajectories.
The core challenge remains accurate estimation of the underlying \textit{flow field}—a problem that has spurred innovations in autoregressive modeling~\cite{flowar}, multi-resolution training~\cite{dimr}, and generalized interpolants design~\cite{nf}.
In essence, flow matching can be viewed as a generalization of the diffusion paradigm:
instead of gradually adding and removing Gaussian noise, it directly learns the continuous trajectories that transport data across domains.
This perspective unifies a family of deterministic generative models and opens new directions for efficient and controllable generation.

ModaFlow builds upon this foundation by leveraging a deterministic flow-matching framework to achieve efficient, stable, and geometrically consistent virtual try-on synthesis.

\subsection{Guidance in Models}

Guidance mechanisms have become a central component in modern diffusion and flow-matching models, enabling control over the trade-off between image fidelity and sample diversity.
The concept was first introduced as \textit{classifier guidance}~\cite{cg}, which leverages gradients from an external classifier to steer the sampling trajectory toward desired conditions.
Subsequently, \textit{classifier-free guidance} (CFG)~\cite{cfg} eliminated the need for a separate classifier by jointly training conditional and unconditional branches, greatly enhancing controllability and practicality.

A series of studies have further refined CFG to address its theoretical and practical limitations.
CFG++~\cite{cfg++} constrains guidance within the data manifold to mitigate off-manifold artifacts and mode collapse.
Analysis of guidance weight schedulers~\cite{weightschedulers} demonstrates that dynamically adjusting guidance strength can improve generation consistency, while AutoG~\cite{autoG} introduces self-guidance using a weaker model instance to enhance quality without sacrificing diversity.
Recent works such as ReCFG~\cite{recfg} and Characteristic Guidance~\cite{charguid} revisit the mathematical formulation of CFG, providing rectified or nonlinear corrections that stabilize sampling under large guidance scales.

Building upon this line of work, CFG-Zero*~\cite{cfgzero} proposes an adaptive guidance weighting scheme and a zero-initialized guidance formulation specifically designed for flow-matching models.

However, the application of such advanced guidance strategies in multimodal conditional generation—where visual and textual priors coexist—remains largely unexplored.
ModaFlow introduces a modality-aware guidance scheme that asymmetrically combines deterministic visual guidance and adaptive textual control via CFG-Zero*, enabling precise text–garment alignment without sacrificing stability.

\section{Preliminary}
In this section, we briefly revisit the foundations of diffusion and flow-matching generative models, which form the backbone of our proposed framework.

\paragraph{Diffusion Models.}
Denoising Diffusion Probabilistic Models (DDPMs)~\cite{ddpm} formulate data generation as the reverse of a stochastic noising process.
Starting from a Gaussian prior, the model progressively denoises latent variables to recover the data distribution.
The training objective can be expressed as a denoising loss:
\begin{equation}
    \label{eq:diffusion_obj}
    \mathcal{L}_{\mathrm{diff}}
    = \mathbb{E}_{\,t,\;\mathbf{x}_0,\;\boldsymbol{\epsilon}}
    \left\|
    \boldsymbol{\epsilon}
    - \boldsymbol{\epsilon}_{\theta}(\mathbf{x}_t)
    \right\|^{2},
\end{equation}
where $\mathbf{x}_t = \sqrt{\bar{\alpha}_t} \mathbf{x}_0 + \sqrt{1 - \bar{\alpha}_t} \boldsymbol{\epsilon}$ is a noisy sample at timestep $t$, and $\boldsymbol{\epsilon}_\theta$ predicts the added noise $\boldsymbol{\epsilon}$.
This training objective enables stable optimization and high-fidelity image synthesis, albeit at the cost of a relatively slow multi-step sampling process — a limitation later addressed by deterministic flow matching formulations.

\paragraph{Flow Matching.}
Flow Matching (FM)~\cite{fm,recflow,nf} provides a deterministic alternative to diffusion by directly learning a continuous-time velocity field $v_{\theta}(\mathbf{x}, t)$ that transports samples from a noise distribution to the data manifold through an ordinary differential equation (ODE).
The corresponding training objective regresses the model velocity to the ground-truth conditional velocity field:
\begin{equation}
    \label{eq:flow_obj}
    \mathcal{L}_{\mathrm{CFM}}
    = \mathbb{E}_{t,\;\mathbf{y},\;\mathbf{x}_{t}\sim p_{t}(\cdot\mid\mathbf{y})}
    \Big[
    \big\|
    v_{\theta}\bigl(\mathbf{x}_{t};\,\mathbf{y}\bigr)
    - u\bigl(\mathbf{x}_{t};\,\mathbf{y}\bigr)
    \big\|^2
    \Big],
\end{equation}
where $\mathbf{x}_t = (1-t)\mathbf{x}_0 + t\mathbf{x}_1$ and $u(\mathbf{x}_t;\mathbf{y})$ denotes the target velocity.
This formulation allows few-step or even single-step sampling while maintaining generation quality.
Accurately estimating the flow field is therefore key to achieving both visual fidelity and geometric consistency in conditional tasks such as virtual try-on.

\section{ModaFlow}
This section details our proposed \textit{ModaFlow} model, a flow-matching based framework designed for high-fidelity virtual try-on with precise text-garment alignment.
As illustrated in Figure~\ref{fig:figpipe}, our model takes as input a \textit{triptych} \([\mathbf{g}, \mathbf{p}, \mathbf{x}_{\text{agn}}]\) — consisting of a reference garment image $\mathbf{g}$, an OpenPose~\cite{openpose} map $\mathbf{p}$, and a garment agnostic person image $\mathbf{x}_{\text{agn}}$ obtained by applying a mask to the person image — along with another auxiliary mask $\mathbf{m}$ that specifies the target garment region.

The model aims to reconstruct a triptych $[\hat{\mathbf{g}}, \hat{\mathbf{p}}, \hat{\mathbf{x}}]$ that matches the ground-truth $[\mathbf{g}, \mathbf{p}, \mathbf{x}_{\text{gt}}]$, thereby learning a consistent flow field across all components.

\textit{ModaFlow} integrates two distinct guidance streams:
a \textit{visual garment embedding} $\mathbf{e}_v$, extracted from \(\mathbf{g}\) using a pretrained image-prompt adapter, and a \textit{textual embedding} $\mathbf{e}_t$, derived from a garment description.
We combine them via our modality-aware guidance design (Section~\ref{subsec:moda-aware}), and further improve geometric consistency with flow-field regularization (Section~\ref{subsec:flow reg}).
The auxiliary region mask $\mathbf{m}$ is obtained by our mask manipulation scheme (Section~\ref{subsec:mask}) to improve robustness under diverse occlusions.

We adopt a flow-matching backbone (implemented with FLUX.1-Fill-dev~\cite{flux}), which jointly denoises the triptych under the influence of both visual and textual guidance.

\subsection{Modality-Aware Guidance Design}
\label{subsec:moda-aware}
In virtual try-on, visual conditions provide precise \emph{structure} (\eg, geometry, texture), while text primarily specifies \emph{semantics} (\eg, style/attributes). Treating them symmetrically often entangles these signals and harms either alignment or detail fidelity.
In order to condition the try-on process on both garment appearance and semantic description, our framework asymmetrically treats visual and textual conditions to reflect their distinct roles in virtual try-on.

\textbf{Visual Garment Embedding.} 
Flux-Redux extracts the visual garment embedding $\mathbf{e}_v$ from the reference image $\mathbf{g}$.
This full-time visual guidance is injected into the backbone at every timestep without stochastic masking, ensuring consistent structural alignment between the input garment and the synthesized output.
This design leverages the fact that the visual condition already provides precise spatial and textural information and thus benefits the fine-grained preservation of logos, patterns, and fabric appearance throughout the generation process.

\textbf{Textual Garment Embedding.}
To guide the model with semantic garment attributes, we generate a natural-language prompt of the target garment using an off-the-shelf captioner~\cite{qwen3} (offline and cached) and encode it into a textual embedding \(\mathbf{e}_t\).
This prompt describes high-level characteristics such as brand, style (\eg, ``boxy'', ``fitted''), material (\eg., ``denim'', ``silk''), and structural details (\eg, ``long sleeves'', ``collarless'').
With this semantic prior, the model can better preserve garment geometry and global coherence, especially in regions where the visual input is occluded or ambiguous.

The two guidance embeddings are thus defined as:
\[
\mathbf{e}_v = f_{\mathrm{redux}}\bigl(\mathbf{g}\bigr), \;\;
\mathbf{e}_t = f_{\mathrm{text}}\bigl(\mathbf{c}\bigr).
\]
where \(\mathbf{c}\) denotes a garment description.
We apply \textit{complementary fusion} where \(\mathbf{e}_{v}\) and \(\mathbf{e}_{t}\) occupy disjoint subspaces through zero-padding, yielding a unified conditioning vector \(\mathbf{e}_{f}\) that preserves both visual structure and semantic attributes without interference.

\textbf{Classifier-Free Guidance.}
To enhance robustness against incomplete or noisy visual conditions, we apply classifier-free guidance (CFG) exclusively to the textual embedding \(\mathbf{e}_{t}\).
During training, the text prompt is randomly dropped with probability \(p\), enabling the model to learn a unified generation process that functions conditionally and unconditionally.

In inference, we adopt the CFG-Zero*~\cite{cfgzero} formulation.
The guided velocity field at step \(t\) is formulated as:
\begin{equation}
    \label{eq:cfg}
    \tilde v_{\theta}\bigl(\mathbf{x}_{t};\mathbf{e}_v,\mathbf{e}_t\bigr)
    = (1 - \omega)\cdot s^{*} \cdot v_{\theta}\bigl(\mathbf{x}_{t};\mathbf{e}_v\bigr)
    \,+\,\omega \cdot v_{\theta}\bigl(\mathbf{x}_{t};\mathbf{e}_f\bigr)\,.
\end{equation}
where the optimized scale \(s^{*}\) is computed by projecting the conditional velocity onto the unconditional velocity:
\begin{equation}
    \label{eq:scales}
    s^{*}
    = \frac{
    v_{\theta}\bigl(\mathbf{x}_{t};\,\mathbf{e}_f\bigr)^{\!\top}\;
    v_{\theta}\bigl(\mathbf{x}_{t};\,\mathbf{e}_v\bigr)
    }{
    \left\|\,v_{\theta}\bigl(\mathbf{x}_{t};\,\mathbf{e}_v\bigr)\right\|^{2}
    }.
\end{equation}
In addition, a zero-init strategy is applied which zeros out the velocity for the first few steps to stabilize sampling in the early stage.

This modality-aware strategy allows \textit{ModaFlow} to preserve pose and garment structure through deterministic visual guidance, while precisely aligning textual guidance to recover plausible garment details and activating semantic descriptions only when beneficial.

\subsection{Flow Field Regularization}
\label{subsec:flow reg}

To improve the directional consistency and perceptual realism of the learned flow field \(v_{\theta}(\mathbf{x}_t)\), we introduce two auxiliary regularization losses that promote both geometric alignment and perceptual fidelity of the predicted flow.

\textbf{Cosine Similarity Loss.}
We penalize the angular discrepancy between the predicted velocity field \(v_\theta(\mathbf{x}_t)\) and the ground-truth target velocity \(u(\mathbf{x}_t)\):
\begin{equation}
    \label{eq:cos}
    \mathcal{L}_{\mathrm{COS}}
    = 1 \,-\, 
    \frac{\,v_{\theta}(\mathbf{x}_t)^{\top}\;u(\mathbf{x}_t)\,}
    {\;\left\|\,v_{\theta}(\mathbf{x}_t)\right\|\;\left\|u(\mathbf{x}_t)\right\|\,}.
\end{equation}

This loss encourages directionally consistent flows between the network output and the optimal trajectory, thus improving the alignment of garment deformation with the underlying person pose and body structure.

\textbf{Perceptual Flow Regularization.}
We adopt a two-stage strategy to enhance the perceptual quality of the predicted velocity field. Since point-wise similarity is insufficient to capture realistic velocity patterns, we learn a feature space for flows and regularize in that space.
First, we pre-train a lightweight Transformer-based flow encoder \(E_\psi\) to discriminate between ground-truth and model-predicted flows.
Given a ground-truth flow \(\mathbf{v}_{gt}\) and a predicted flow \(\mathbf{v}_{pred}\) at timestep \(t\), the encoder is trained to produce discriminative flow features that separate predicted flows from ground-truth flows:
\begin{equation}
    \mathcal{L}_{\mathrm{perc}}^{\mathrm{pretrain}}
    = \; - \; \Big\|\, E_{\psi}\bigl(\mathbf{v}_{pred},\,t\bigr)
    \;-\; E_{\psi}\bigl(\mathbf{v}_{gt},\,t\bigr)\Big\|^{2}.
\end{equation}

After pre-training, $E_\psi$ is frozen and repurposed as a perceptual regularizer.
During training, we minimize the feature distance between predicted and ground-truth flows:
\begin{equation}
    \label{eq:perception}
    \mathcal{L}_{\mathrm{perc}}
    = \Big\|\, E_{\psi}\bigl(\mathbf{v}_{pred},\,t\bigr)
           - E_{\psi}\bigl(\mathbf{v}_{gt},\,t\bigr)\Big\|^{2}.
\end{equation}

This design ensures that the flow encoder captures discriminative features of realistic velocity fields, which in turn guide the generator to produce flows that are perceptually aligned with the true data distribution, stabilizing correspondence and reduces texture/shape artifacts. The combined training objective thus becomes the following.
\begin{equation}
    \label{eq:total_loss}
    \mathcal{L}_{\mathrm{total}}
    = \mathcal{L}_{\mathrm{CFM}}
      + \lambda_{\mathrm{COS}}\;\mathcal{L}_{\mathrm{COS}}
      + \lambda_{\mathrm{perc}}\;\mathcal{L}_{\mathrm{perc}},
\end{equation}
where \(\mathcal{L}_{\mathrm{CFM}}\) denotes the conditional flow matching loss, and \(\lambda_{\mathrm{COS}}, \lambda_{\mathrm{perc}}\) are weighting hyper-parameters.

Together, these regularizers ensure that the learned vector field is not only mathematically consistent but also perceptually plausible, improving garment alignment and reducing artifacts in virtual try-on.

\subsection{Mask Manipulation and Strategy}
\label{subsec:mask}

In practical unpaired try-on, the target garment often differs in coverage from the source outfit (\eg, short-sleeve to  long-sleeve), making tight masks mismatched and overly restrictive.
To improve model robustness to varying levels of body-garment occlusion and to simulate real-world unpaired try-on conditions, we design a mask manipulation scheme that dynamically varies the masking pattern during training.

\textbf{Mask Types.}  
We define three main mask types to acquire the garment-agnostic person image input \(\mathbf{x}_{\text{agn}}\):

\begin{itemize}
\item \textbf{Box mask}: a tight bounding box around the garment region, providing minimal body context, mimicking coarse localization;
\item \textbf{Relaxed mask}: a mask region dilated beyond the garment boundary, intentionally covering the adjacent context to improve the model's ability to reason about incomplete structural cues;
\item \textbf{Transparent mask}: a semi-transparent overlay of the same box mask shape, which retains partial visual information of the original garment, providing soft geometry cues under occlusion and teaching the model to infer missing structure. We empirically found that a transparency level of 10\% provides a good trade-off. We observed that higher levels can lead to overfitting.
\end{itemize}

\begin{wrapfigure}[19]{r}{0.50\textwidth}
  \vspace{-2pt}
  \centering
  \begin{subfigure}{\linewidth}
    \centering
    \includegraphics[width=\linewidth]{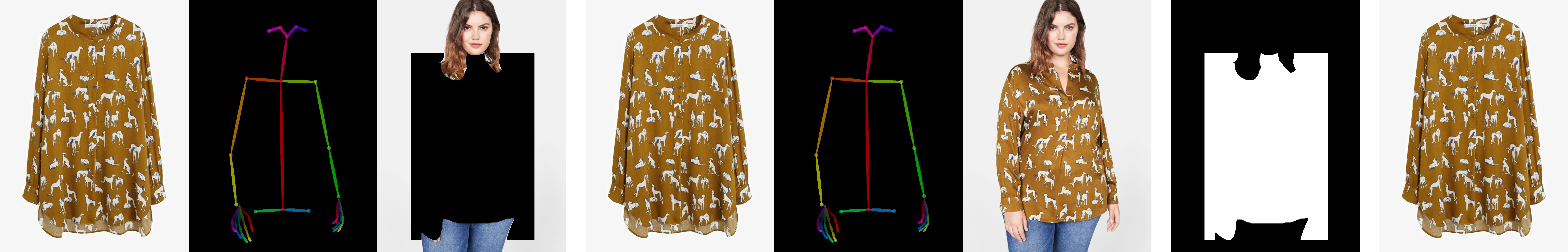}
    \caption{Training batch with box masks.}
    \label{fig:top}
  \end{subfigure}
  
  \begin{subfigure}{\linewidth}
    \centering
    \includegraphics[width=\linewidth]{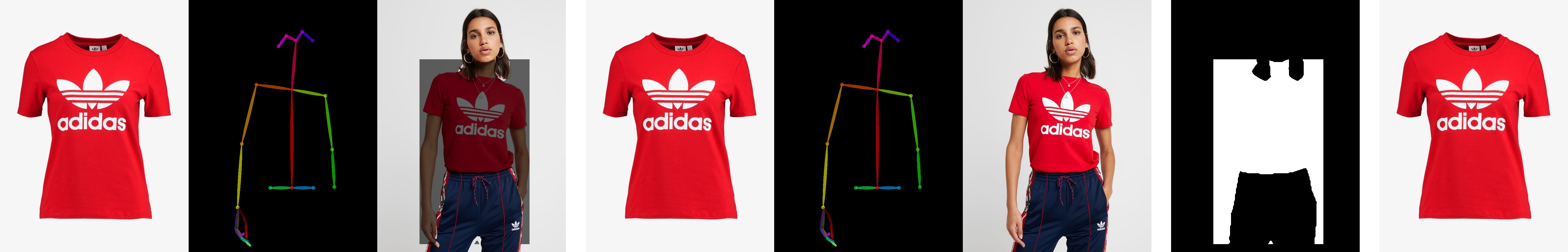}
    \caption{Training batch with transparent masks.}
    \label{fig:middle}
  \end{subfigure}
  
  \begin{subfigure}{\linewidth}
    \centering
    \includegraphics[width=\linewidth]{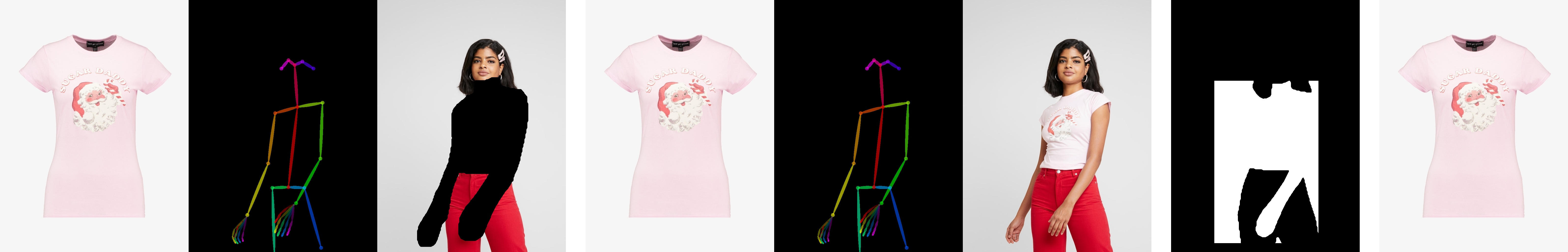}
    \caption{Training batch with relaxed masks.}
    \label{fig:bottom}
  \end{subfigure}

  \caption{Three mask types used during training to improve robustness under diverse occlusion conditions. From left to right in each batch: triptych input, triptych ground-truth, auxiliary mask and garment. For clearer visualization, the transparency level in (b) is set to 40\% instead of 10\% used in training.}
  \label{fig:batch}
\end{wrapfigure}

We show an example of a training batch in Figure~\ref{fig:batch}, illustrating the three masking regimes.

\textbf{Mask Sampling Strategy.}  
During training, for each sample triptych \([\mathbf{g}, \mathbf{p}, \mathbf{x}_{\text{agn}}]\), we stochastically assign one of the mask types to acquire \(\mathbf{x}_{\text{agn}}\) according to the distribution: 90\% box mask, 5\% transparent mask, and 5\% relaxed mask. 
This sampling prioritizes the box-mask regime to match practical unpaired try-on with only coarse garment localization, while transparent/relaxed masks provide occasional geometry/context cues only as a regularizer; we found the performance \emph{not overly sensitive} to nearby ratios.
For the auxiliary mask \(\mathbf{m}\) that indicates the target region for generation, we always use the box mask configuration to align with the unpaired virtual try-on setting. 
This design acts as a dropout-like regularization: under box masks, the model relies solely on the reference garment to complete missing regions and infer plausible body structure; under transparent masks, it can leverage partial body evidence to stabilize geometry/pattern.

This strategy guarantees that \textit{ModaFlow} is exposed to diverse masking conditions during training, enabling it to infer convincing body structures from the garment and pose alone when only a box mask is available at inference time.
\FloatBarrier
\section{Experiments}
\label{sec:experiments}

We evaluate \textit{ModaFlow} on paired and unpaired virtual try-on tasks using VITON-HD~\cite{vitonhd} and DressCode~\cite{dresscode} datasets.
Experiments include comparisons with state-of-the-art methods, detailed quantitative and qualitative analyses, and ablation studies.

\subsection{Datasets and Implementation Details}
\label{subsec:datasets}

We use two high-resolution virtual try-on benchmarks: \textbf{VITON-HD}~\cite{vitonhd} and \textbf{DressCode}~\cite{dresscode}. 
VITON-HD contains 11{,}647 training and 2{,}032 testing image pairs of resolution $1024\times768$, while DressCode includes 13{,}563 training and 1{,}800 testing samples with the same resolution.
Both datasets provide paired images of a person and an in-shop garment; we use the official test splits for paired and unpaired evaluation sets. In unpaired evaluation, we randomly reassign garments across subjects to assess the model’s generalization under real-world conditions.

\textit{ModaFlow} is implemented on top of the \textit{InsertAnything}~\cite{insertanything} configuration for the \textit{FLUX.1-Fill-dev} backbone~\cite{flux}, while the \textit{FLUX.1-Redux-dev} image prompt adapter is kept frozen.
In this paper, we rely on an off-the-shelf image captioner, \textit{Qwen3-VL-30B-A3B-Instruct}~\cite{qwen3}, to obtain a natural-language garment description which we perform and cache it once offline. However, \textit{ModaFlow} is not dependent on the captioner and it 
can be replaced by any alternative caption source (\eg, another captioner or manually written prompts). We consider that the evaluation of the captioner quality/noise is out of the scope of this paper.
\textit{ModaFlow} is fine-tuned using LoRA~\cite{lora} with a rank of 256, and optimized with the \textit{Prodigy} optimizer~\cite{prodigy}. 
We set the CFG dropout probability $p=0.1$. 
Training is performed on two NVIDIA H100 GPUs. 
To enable the perceptual flow loss, we pre-train a lightweight Transformer-based flow encoder on 40{,}000 sampled pairs of real and predicted flows, each annotated with their corresponding timestep.
Separate models are trained for VITON-HD and DressCode to respect dataset-specific characteristics. All models are trained at a resolution of $1024\times768$ and evaluated on both paired and unpaired settings for all datasets.

\begin{table}[t]
\setlength{\extrarowheight}{-7pt}
\centering
\caption{Quantitative comparison with state-of-the-art methods on DressCode and VITON-HD datasets under paired and unpaired settings. 
$\uparrow$ indicates higher is better, $\downarrow$ indicates lower is better.}
\label{tab:quantitative_comparison}
\resizebox{\textwidth}{!}{
\begin{tabular}{l|cccc|cc|cccc|cc}
\toprule
\multirow{3}{*}{Methods} 
& \multicolumn{6}{c|}{VITON-HD} 
& \multicolumn{6}{c}{DressCode} \\
\cmidrule(lr){2-13}
& \multicolumn{4}{c|}{Paired} 
& \multicolumn{2}{c|}{Unpaired} 
& \multicolumn{4}{c|}{Paired} 
& \multicolumn{2}{c}{Unpaired} \\
\cmidrule(lr){2-13}

& SSIM$\uparrow$ & LPIPS$\downarrow$ & FID$\downarrow$ & KID$\downarrow$ & FID$\downarrow$ & KID$\downarrow$
& SSIM$\uparrow$ & LPIPS$\downarrow$ & FID$\downarrow$ & KID$\downarrow$ & FID$\downarrow$ & KID$\downarrow$ \\
\midrule
IDM-VTON~\cite{idmvton}&0.846 &0.083 &9.074 &1.357 &12.732 &1.677 &0.871 &0.067 &7.182 &1.818 &9.123 &2.446\\
StableVTON~\cite{stableviton}&0.851 &0.079 &8.726 &1.273 &11.969 &1.555 &0.876 &0.063 &6.691 &1.663 &8.965 &2.386\\
TPD~\cite{tpd}&0.880 &0.060 &7.028 &0.901 &9.536 &1.212 &0.905 &0.045 &5.124 &1.258 &8.174 &1.928\\
Any2AnyTryOn~\cite{any2anytryon}&0.869&0.069&7.614&1.231&11.804&1.476&0.883&0.053&5.417&1.340&8.041&2.163\\
CatVTON~\cite{catvton}&0.885 &0.062 &6.43 &0.841 &9.418 &0.983 &0.910 &0.039 &4.780 &0.963 &7.317 &1.713\\
FitDiT~\cite{fitdit}&0.890 &0.058 &6.583 &0.619 &9.273 &0.851 &0.915 &0.038 &4.334 &0.919 &6.756 &1.570\\
\midrule
\textbf{ModaFlow}&\textbf{0.905} &\textbf{0.053} &\textbf{5.282} &\textbf{0.324} &\textbf{8.041} &\textbf{0.637} &\textbf{0.930} &\textbf{0.031} &\textbf{3.210} &\textbf{0.694} &\textbf{6.011} &\textbf{1.317}\\
\bottomrule
\end{tabular}}
\end{table}

\subsection{Evaluation Metrics}
\label{subsec:metrics}

We adopt standard image synthesis metrics to quantitatively evaluate both paired and unpaired settings. 
For paired evaluation, where ground-truth try-on images are available, we measure:
\textbf{SSIM}~\cite{ssim} for structural similarity, \textbf{LPIPS}~\cite{lpips} for perceptual distance based on deep feature embeddings, and distributional measures including \textbf{FID}~\cite{fid} and \textbf{KID}~\cite{kid} to assess realism and diversity. 
In contrast, for unpaired evaluation, the direct reference‐based metrics such as SSIM and LPIPS are not applicable; hence we rely solely on the distribution‐based metrics FID and KID.

\subsection{Comparison with State-of-the-Art Methods}
\label{subsec:sota}

\textbf{Quantitative Results.}
As shown in Table~\ref{tab:quantitative_comparison}, ModaFlow achieves the best performance across all evaluation metrics.
We additionally report inference cost in Table~\ref{tab:cost}, including a simplified ModaFlow variant (w/o CFG and fewer steps).
Our method yields significant improvements in FID and KID, confirming its superior image realism and distributional fidelity.
The higher SSIM and lower LPIPS scores demonstrate that ModaFlow preserves structural integrity and fine garment details more effectively.
These improvements can be attributed to the proposed flow field regularization, modality-aware guidance, and mask manipulation strategy that collectively enhance geometric consistency and semantic alignment.

\begin{figure}[t]
  \centering
  \includegraphics[width=\linewidth]{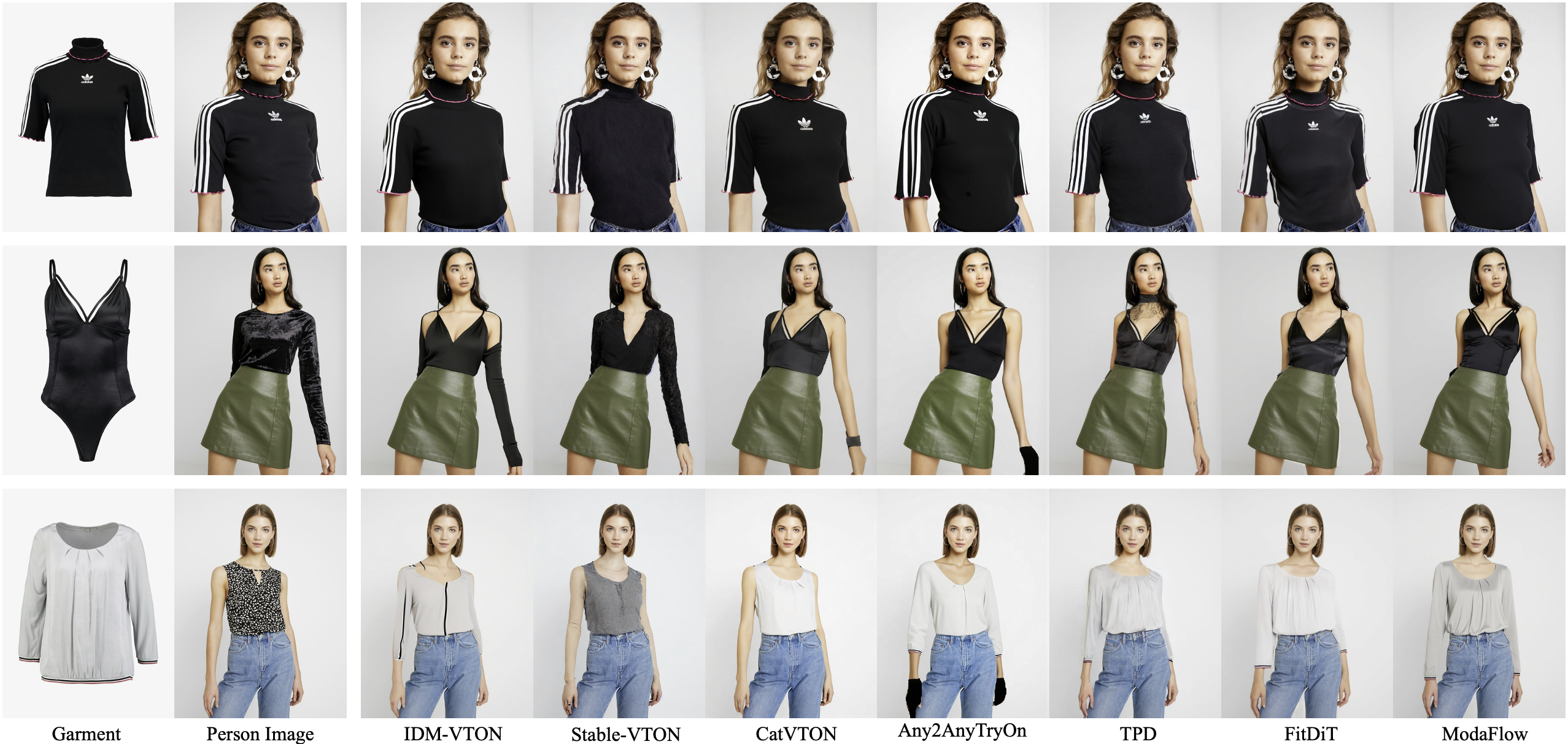}
  \caption{Qualitative comparison on the \textbf{VITON-HD} dataset.
The first row shows paired results, while the second and third rows show \textbf{unpaired cross-category} try-on cases (\eg short-sleeve to long-sleeve).
ModaFlow produces photo-realistic results with consistent body alignment and fine garment details across all settings.}
  \label{fig:vthd}
  \vspace{-13pt}
\end{figure}

\begin{wrapfigure}{r}{0.5\textwidth}
  \vspace{-25pt}
  \centering
  \includegraphics[width=\linewidth]{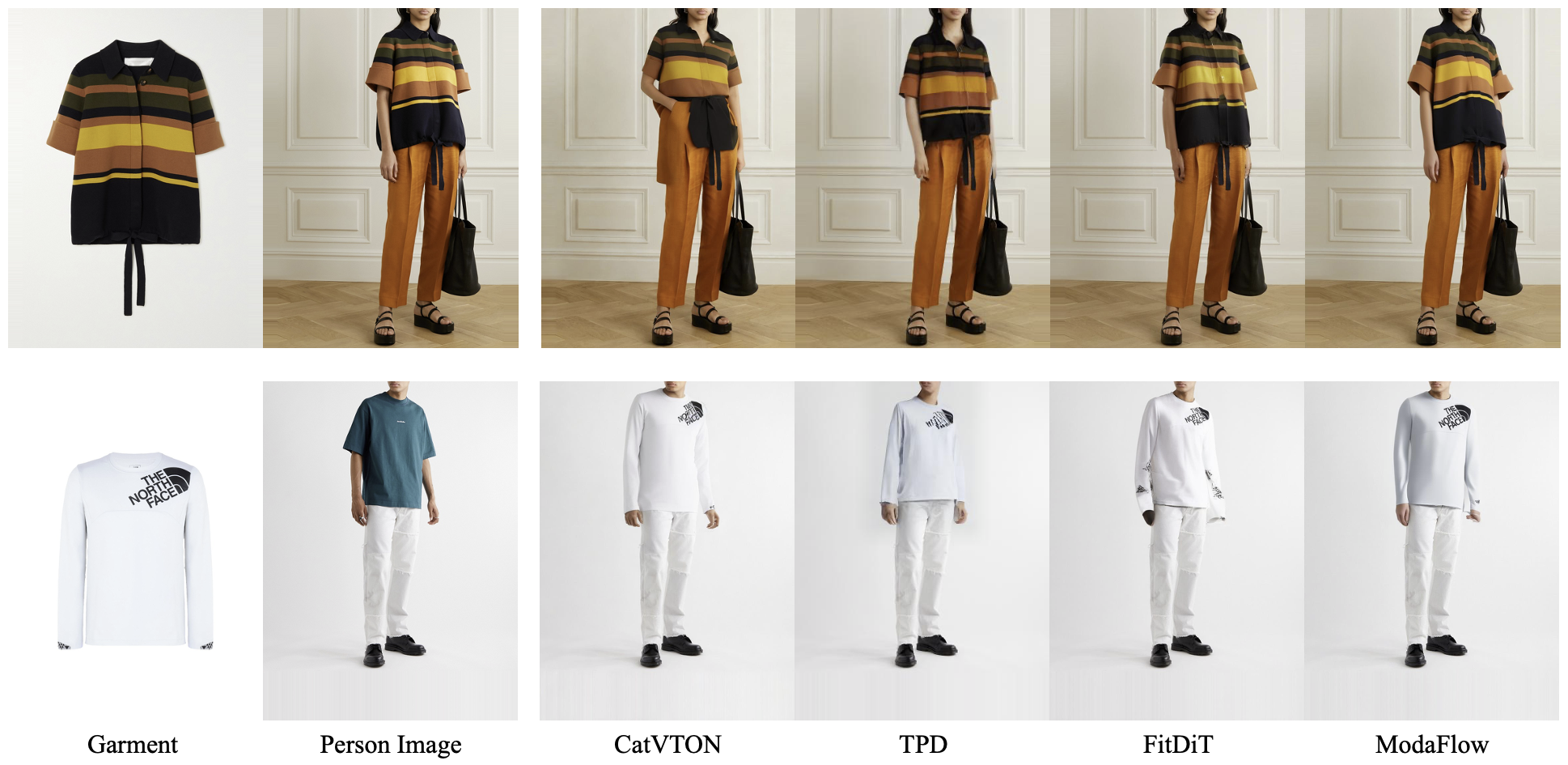}
  \caption{Qualitative comparison on the \textbf{DressCode} dataset. 
The top row shows paired and the bottom row unpaired settings. 
ModaFlow maintains garment fidelity and pose consistency across diverse categories and lighting conditions, outperforming previous approaches in overall realism.}
  \label{fig:dc}
  \vspace{-5pt}
\end{wrapfigure}

\textbf{Qualitative Results.}
Figure~\ref{fig:vthd} and Figure~\ref{fig:dc} showcase the results on the \textbf{VITON-HD} and \textbf{DressCode} datasets, illustrating the qualitative comparison in paired and unpaired settings. We further present \textbf{in-the-wild} try-on results in Figure~\ref{fig:wild}, demonstrating robustness under unconstrained backgrounds, lighting, and poses. ModaFlow not only preserves garment semantics (texture, pattern, logo), but also adapts to diverse body geometries and poses, even in unpaired or cross-category scenarios.
In contrast, prior methods exhibit texture smoothing or body misalignment, particularly in challenging cross-category try-on cases. 
Notably, in unpaired trials where garments are transferred across different identities and body types, ModaFlow exhibits minimal distortion and retains garment-body alignment.


Overall, the qualitative results demonstrate that ModaFlow achieves superior fidelity and generalization across datasets, garment types, and body geometries, bridging the gap between realistic visual synthesis and geometric adaptability in image-based virtual try-on.


\subsection{Ablation Study}
\label{subsec:ablation}

\begin{figure}[t]
\centering
\begin{minipage}[t]{0.49\textwidth}
  \centering
  \includegraphics[width=\linewidth]{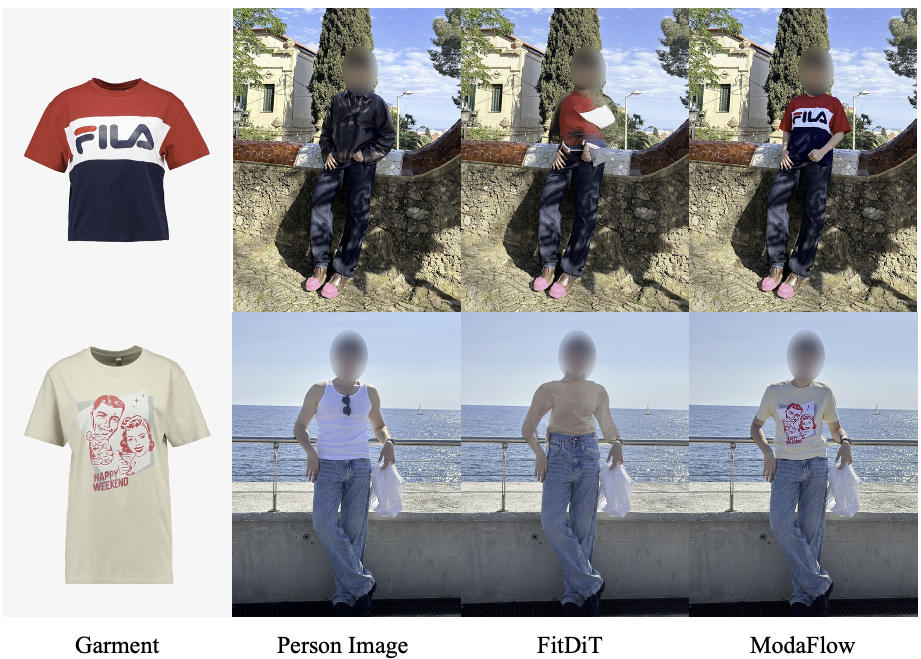}
  \caption{In-the-wild try-on results. ModaFlow synthesizes plausible try-on outputs with better garment fidelity and alignment.}
  \label{fig:wild}
\end{minipage}\hfill
\begin{minipage}[t]{0.49\textwidth}
  \centering
  \includegraphics[width=\linewidth]{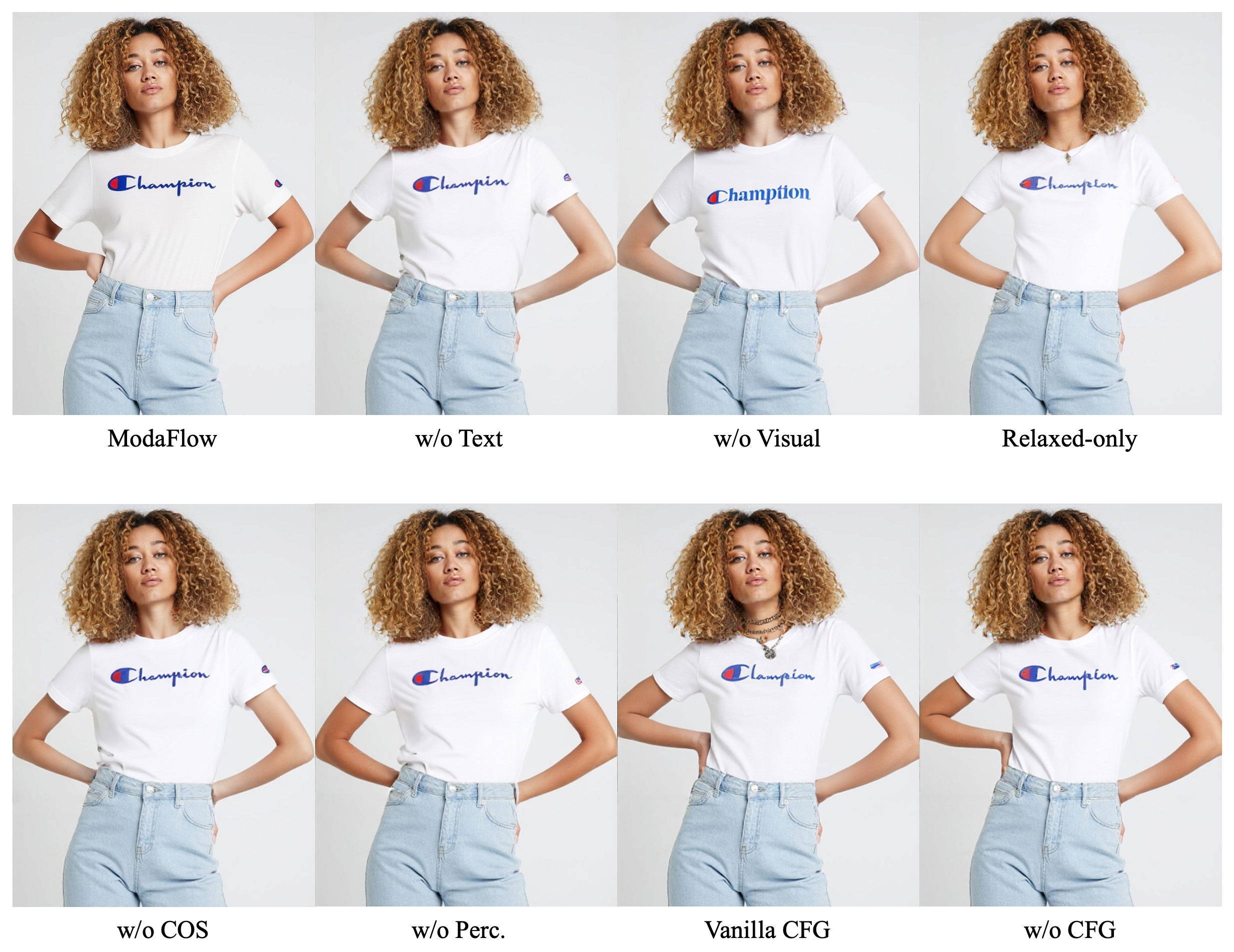}
  \caption{Visual ablation on VITON-HD. Removing mask manipulation or flow regularization causes blur/misalignment; full ModaFlow is sharper and more consistent.}
  \label{fig:ablation}
\end{minipage}
\end{figure}

We perform ablation studies on the VITON-HD dataset to assess the contribution of each key component in ModaFlow, including modality-aware guidance, mask manipulation, flow field regularization, and CFG-Zero*.
Quantitative results are reported in Table~\ref{tab:ablation}, with qualitative comparisons in Figure~\ref{fig:ablation}.

\textbf{Effect of Modality-Aware Guidance.}
We first analyze the impact of the proposed asymmetric conditioning scheme.
Removing the textual guidance (\textit{w/o Text}) leads to a noticeable degradation in FID and LPIPS, and weaker alignment between garment semantics and the generated appearance.
Removing the visual guidance (\textit{w/o Visual}) severely corrupts geometry and texture fidelity, confirming that full-time visual conditioning is essential, while the text branch acts as a complementary semantic controller.
We also observe non-trivial text controllability under a fixed visual reference (see the suppl. material).

\textbf{Effect of Mask Manipulation.}
To validate our mask strategy, we compare ModaFlow with a variant trained using only relaxed masks (\textit{Relaxed Only}).
While the paired performance remains similar, unpaired FID/KID drop significantly and visual results exhibit garment drifting and body–cloth misalignment in Figure~\ref{fig:ablation}.
This confirms that training with box masks aligned to the inference setup, together with stochastic mask sampling, is crucial for robust unpaired generalization.

\textbf{Effect of Flow Regularization.}
We evaluate variants without cosine similarity loss (\textit{w/o Cos}), without perceptual loss (\textit{w/o Perc}), and without both (\textit{w/o Both}).
All three variants perform worse than the full model, with \textit{w/o Both} showing the largest increase in LPIPS and FID.
This demonstrates that the two regularizers are complementary: cosine loss constrains the velocity direction, while perceptual flow regularization encourages realistic flow patterns in feature space.

\textbf{Effect of CFG-Zero*.}
Finally, we compare standard classifier-free guidance (\textit{Vanilla CFG}) with our CFG-Zero* based sampler.
Without CFG, the model already produces stable and visually consistent results, but tends to under-utilize semantic cues from the textual embedding, occasionally leading to weaker alignment in style or attribute details.
Introducing CFG-Zero* yields moderate improvements in global garment coherence and text–appearance alignment, while avoiding the artifacts and geometric instability observed with vanilla CFG at large guidance scales.

\begin{table}[t]
\centering
\begin{minipage}[t]{0.46\textwidth}
\centering
\caption{Inference cost comparison. We report sampling steps, inference time and peak GPU memory under default settings. We also report a simplified ModaFlow variant (w/o CFG and fewer steps) as an efficiency-oriented setting.}
\label{tab:cost}
\resizebox{\linewidth}{!}{%
\begin{tabular}{lccccccc}
\toprule
Method&Steps&Inference Time(s)&GPU Memory(M)\\
\midrule
IDM-VTON &30&13.29&25,184\\
StableVTON &50&79.16&20,052\\
TPD &50&23.72&30,017 \\
CatVTON & 50 & 12.70 & 5,940\\
FitDiT & 25 &4.57&19,950 \\
Any2AnyTryOn & 30 & 65.27 & 36,221\\
\midrule
\textbf{ModaFlow w/o CFG} & 30 & 19.34 & 37,436\\
\textbf{ModaFlow} & 50 & 85.03 & 38,954\\
\bottomrule
\end{tabular}}
\end{minipage}\hfill
\begin{minipage}[t]{0.52\textwidth}
\centering
\caption{Ablation study of ModaFlow on VITON-HD. We report paired SSIM/LPIPS and unpaired FID/KID.}
\label{tab:ablation}
\resizebox{\linewidth}{!}{
\begin{tabular}{lcccccc}
\toprule
\multirow{2}{*}{Variants}
& \multicolumn{4}{c}{Paired}
& \multicolumn{2}{c}{Unpaired}\\
\cmidrule(lr){2-7}
& SSIM$\uparrow$ & LPIPS$\downarrow$ & FID$\downarrow$ & KID$\downarrow$ & FID$\downarrow$ & KID$\downarrow$ \\
\midrule
ModaFlow&\textbf{0.905} &\textbf{0.053} &\textbf{5.282} &\textbf{0.324} &\textbf{8.041} &\textbf{0.637}  \\
w/o Text  &0.873 &0.065 &6.401 &0.795 &9.799 &0.881\\
w/o Visual  &0.871 &0.072 &6.308 &0.802 &9.972 &0.989\\
Relaxed Only &0.875 &0.066 &6.207 &0.784 &9.626 &0.917\\
w/o Cos  &0.877 &0.063 &6.115 &0.748 &9.535 &0.884\\
w/o Perc.  &0.873 &0.069 &6.315 &0.804 &9.823 &0.965\\
w/o Both  &0.870 &0.073 &6.331 &0.813 &10.015 &0.997\\
Vanilla CFG  &0.880 &0.061 &5.911 &0.726 &9.318 &0.862\\
w/o CFG  &0.901 &0.059 &5.393 &0.426 &8.473 &0.812\\
\bottomrule
\end{tabular}}
\end{minipage}
\end{table}
\FloatBarrier
\section{Conclusion}
\label{sec:conclusion}
\vspace{-10pt}
We presented \textit{ModaFlow}, a flow-matching based framework for high-fidelity image-based virtual try-on. 
By introducing a modality-aware guidance mechanism that asymmetrically integrates deterministic visual conditioning and adaptive textual control, \textit{ModaFlow} achieves precise garment–pose alignment and fine-grained semantic consistency. 
The proposed mask manipulation strategy and flow field regularization, comprising cosine similarity and perceptual flow discrimination, jointly improve the model’s robustness to occlusions and improve the directional and perceptual quality of the learned velocity field in unpaired settings. Evaluation shows that \textit{ModaFlow} surpasses existing methods in both qualitative and quantitative evaluations, producing photo-realistic, geometry-consistent, and semantically accurate try-on results. Looking ahead, we plan to further accelerate the inference process through efficient flow integration and adaptive step reduction, as well as to enhance the textual controllability. 


%
%
\bibliographystyle{splncs04}
\bibliography{main}

\begin{thebibliography}{10}
\providecommand{\url}[1]{\texttt{#1}}
\providecommand{\urlprefix}{URL }
\providecommand{\doi}[1]{https://doi.org/#1}

\bibitem{nf}
Albergo, M., Vanden-Eijnden, E.: Building normalizing flows with stochastic
  interpolants. In: ICLR 2023 Conference (2023)

\bibitem{kid}
Bi{\'n}kowski, M., Sutherland, D.J., Arbel, M., Gretton, A.: Demystifying mmd
  gans. In: International Conference on Learning Representations (2018)

\bibitem{openpose}
{Cao}, Z., {Hidalgo Martinez}, G., {Simon}, T., {Wei}, S., {Sheikh}, Y.A.:
  Openpose: Realtime multi-person 2d pose estimation using part affinity
  fields. IEEE Transactions on Pattern Analysis and Machine Intelligence
  (2019)

\bibitem{vitonhd}
Choi, S., Park, S., Lee, M., Choo, J.: Viton-hd: High-resolution virtual try-on
  via misalignment-aware normalization. In: Proc. of the IEEE conference on
  computer vision and pattern recognition (CVPR) (2021)

\bibitem{idmvton}
Choi, Y., Kwak, S., Lee, K., Choi, H., Shin, J.: Improving diffusion models for
  authentic virtual try-on in the wild. In: European Conference on Computer
  Vision. pp. 206--235. Springer (2024)

\bibitem{catvton}
Chong, Z., Dong, X., Li, H., Zhang, S., Zhang, W., Zhang, X., Zhao, H., Jiang,
  D., Liang, X.: Catvton: Concatenation is all you need for virtual try-on with
  diffusion models. arXiv preprint arXiv:2407.15886  (2024)

\bibitem{cfg++}
Chung, H., Kim, J., Park, G.Y., Nam, H., Ye, J.C.: {CFG}++:
  Manifold-constrained classifier free guidance for diffusion models. In: The
  Thirteenth International Conference on Learning Representations (2025)

\bibitem{cg}
Dhariwal, P., Nichol, A.: Diffusion models beat gans on image synthesis.
  Advances in neural information processing systems  \textbf{34},  8780--8794
  (2021)

\bibitem{sd3}
Esser, P., Kulal, S., Blattmann, A., Entezari, R., M{\"u}ller, J., Saini, H.,
  Levi, Y., Lorenz, D., Sauer, A., Boesel, F., et~al.: Scaling rectified flow
  transformers for high-resolution image synthesis. In: Forty-first
  international conference on machine learning (2024)

\bibitem{cfgzero}
Fan, W., Zheng, A.Y., Yeh, R.A., Liu, Z.: Cfg-zero*: Improved classifier-free
  guidance for flow matching models. arXiv preprint arXiv:2503.18886  (2025)

\bibitem{gan}
Goodfellow, I.J., Pouget-Abadie, J., Mirza, M., Xu, B., Warde-Farley, D.,
  Ozair, S., Courville, A., Bengio, Y.: Generative adversarial nets. Advances
  in neural information processing systems  \textbf{27} (2014)

\bibitem{any2anytryon}
Guo, H., Zeng, B., Song, Y., Zhang, W., Liu, J., Zhang, C.: Any2anytryon:
  Leveraging adaptive position embeddings for versatile virtual clothing tasks.
  In: Proceedings of the IEEE/CVF International Conference on Computer Vision.
  pp. 19085--19096 (2025)

\bibitem{viton}
Han, X., Wu, Z., Wu, Z., Yu, R., Davis, L.S.: Viton: An image-based virtual
  try-on network. In: Proceedings of the IEEE conference on computer vision and
  pattern recognition. pp. 7543--7552 (2018)

\bibitem{ddpm}
Ho, J., Jain, A., Abbeel, P.: Denoising diffusion probabilistic models.
  Advances in neural information processing systems  \textbf{33},  6840--6851
  (2020)

\bibitem{cfg}
Ho, J., Salimans, T.: Classifier-free diffusion guidance. arXiv preprint
  arXiv:2207.12598  (2022)

\bibitem{lora}
Hu, E.J., Shen, Y., Wallis, P., Allen-Zhu, Z., Li, Y., Wang, S., Wang, L.,
  Chen, W.: Lo{RA}: Low-rank adaptation of large language models. In:
  International Conference on Learning Representations (2022),
  \url{https://openreview.net/forum?id=nZeVKeeFYf9}

\bibitem{fitdit}
Jiang, B., Hu, X., Luo, D., He, Q., Xu, C., Peng, J., Zhang, J., Wang, C., Wu,
  Y., Fu, Y.: Fitdit: Advancing the authentic garment details for high-fidelity
  virtual try-on. CoRR  (2024)

\bibitem{autoG}
Karras, T., Aittala, M., Kynk{\"a}{\"a}nniemi, T., Lehtinen, J., Aila, T.,
  Laine, S.: Guiding a diffusion model with a bad version of itself. Advances
  in Neural Information Processing Systems  \textbf{37},  52996--53021 (2024)

\bibitem{stableviton}
Kim, J., Gu, G., Park, M., Park, S., Choo, J.: Stableviton: Learning semantic
  correspondence with latent diffusion model for virtual try-on. In:
  Proceedings of the IEEE/CVF conference on computer vision and pattern
  recognition. pp. 8176--8185 (2024)

\bibitem{flux}
Labs, B.F.: Flux. \url{https://github.com/black-forest-labs/flux} (2024)

\bibitem{tryongan}
Lewis, K.M., Varadharajan, S., Kemelmacher-Shlizerman, I.: Tryongan: Body-aware
  try-on via layered interpolation. ACM Transactions on Graphics (TOG)
  \textbf{40}(4),  1--10 (2021)

\bibitem{handfit}
Liang, Y., Hu, X., Jiang, B., Luo, D., Peng, X., Wu, K., Xu, C., Han, W., Jin,
  T., Wang, C., et~al.: Vton-handfit: Virtual try-on for arbitrary hand pose
  guided by hand priors embedding. In: Proceedings of the IEEE/CVF Conference
  on Computer Vision and Pattern Recognition. pp. 22616--22626 (2025)

\bibitem{fm}
Lipman, Y., Chen, R.T., Ben-Hamu, H., Nickel, M., Le, M.: Flow matching for
  generative modeling. In: 11th International Conference on Learning
  Representations, ICLR 2023 (2023)

\bibitem{dimr}
Liu, Q., Zeng, Z., He, J., Yu, Q., Shen, X., Chen, L.C.: Alleviating distortion
  in image generation via multi-resolution diffusion models and time-dependent
  layer normalization. Advances in Neural Information Processing Systems
  \textbf{37},  133879--133907 (2024)

\bibitem{recflow}
Liu, X., Gong, C., Liu, Q.: Flow straight and fast: Learning to generate and
  transfer data with rectified flow. In: The Eleventh International Conference
  on Learning Representations (ICLR) (2023)

\bibitem{mcvton}
Luan, J., Li, G., Zhao, L., Xing, W.: Mc-vton: Minimal control virtual try-on
  diffusion transformer. CoRR  (2025)

\bibitem{prodigy}
Mishchenko, K., Defazio, A.: Prodigy: an expeditiously adaptive parameter-free
  learner. In: Proceedings of the 41st International Conference on Machine
  Learning. pp. 35779--35804 (2024)

\bibitem{dresscode}
Morelli, D., Fincato, M., Cornia, M., Landi, F., Cesari, F., Cucchiara, R.:
  {Dress Code: High-Resolution Multi-Category Virtual Try-On}. In: Proceedings
  of the European Conference on Computer Vision (2022)

\bibitem{fid}
Parmar, G., Zhang, R., Zhu, J.Y.: On aliased resizing and surprising subtleties
  in gan evaluation. In: Proceedings of the IEEE/CVF conference on computer
  vision and pattern recognition. pp. 11410--11420 (2022)

\bibitem{dit}
Peebles, W., Xie, S.: Scalable diffusion models with transformers. In:
  Proceedings of the IEEE/CVF international conference on computer vision. pp.
  4195--4205 (2023)

\bibitem{sdxl}
Podell, D., English, Z., Lacey, K., Blattmann, A., Dockhorn, T., M{\"u}ller,
  J., Penna, J., Rombach, R.: Sdxl: Improving latent diffusion models for
  high-resolution image synthesis. In: The Twelfth International Conference on
  Learning Representations (2024)

\bibitem{dreamfusion}
Poole, B., Jain, A., Barron, J.T., Mildenhall, B.: Dreamfusion: Text-to-3d
  using 2d diffusion. In: ICLR (2023)

\bibitem{cit}
Ren, B., Tang, H., Meng, F., Runwei, D., Torr, P.H., Sebe, N.: Cloth
  interactive transformer for virtual try-on. ACM Transactions on Multimedia
  Computing, Communications and Applications  \textbf{20}(4),  1--20 (2023)

\bibitem{flowar}
Ren, S., Yu, Q., He, J., Shen, X., Yuille, A., Chen, L.C.: Flowar: Scale-wise
  autoregressive image generation meets flow matching. In: Proceedings of the
  Forty-second International Conference on Machine Learning (ICML) (2025)

\bibitem{styleganxl}
Sauer, A., Schwarz, K., Geiger, A.: Stylegan-xl: Scaling stylegan to large
  diverse datasets. In: ACM SIGGRAPH 2022 conference proceedings. pp. 1--10
  (2022)

\bibitem{insertanything}
Song, W., Jiang, H., Yang, Z., Quan, R., Yang, Y.: Insert anything: Image
  insertion via in-context editing in dit. arXiv preprint arXiv:2504.15009
  (2025)

\bibitem{dualfit}
Tran, M., Clements, J., Manoharan, A.P., Nguyen, T., Le, N.: Dualfit: A
  two-stage virtual try-on via warping and synthesis. In: Proceedings of the
  IEEE/CVF International Conference on Computer Vision. pp. 2397--2407 (2025)

\bibitem{cpvton}
Wang, B., Zheng, H., Liang, X., Chen, Y., Lin, L., Yang, M.: Toward
  characteristic-preserving image-based virtual try-on network. In: Proceedings
  of the European conference on computer vision (ECCV). pp. 589--604 (2018)

\bibitem{weightschedulers}
Wang, X., Dufour, N., Andreou, N., Cani, M.P., Abrevaya, V.F., Picard, D.,
  Kalogeiton, V.: Analysis of classifier-free guidance weight schedulers.
  Transactions on Machine Learning Research Journal  (2024)

\bibitem{ssim}
Wang, Z., Bovik, A.C., Sheikh, H.R., Simoncelli, E.P.: Image quality
  assessment: from error visibility to structural similarity. IEEE transactions
  on image processing  \textbf{13}(4),  600--612 (2004)

\bibitem{recfg}
Xia, M., Xue, N., Shen, Y., Yi, R., Gong, T., Liu, Y.J.: Rectified diffusion
  guidance for conditional generation. In: Proceedings of the Computer Vision
  and Pattern Recognition Conference. pp. 13371--13380 (2025)

\bibitem{ootd}
Xu, Y., Gu, T., Chen, W., Chen, A.: Ootdiffusion: Outfitting fusion based
  latent diffusion for controllable virtual try-on. In: Proceedings of the AAAI
  Conference on Artificial Intelligence. vol.~39, pp. 8996--9004 (2025)

\bibitem{qwen3}
Yang, A., Li, A., Yang, B., Zhang, B., Hui, B., Zheng, B., Yu, B., Gao, C.,
  Huang, C., Lv, C., et~al.: Qwen3 technical report. arXiv preprint
  arXiv:2505.09388  (2025)

\bibitem{acgpn}
Yang, H., Zhang, R., Guo, X., Liu, W., Zuo, W., Luo, P.: Towards
  photo-realistic virtual try-on by adaptively generating-preserving image
  content. In: Proceedings of the IEEE/CVF conference on computer vision and
  pattern recognition. pp. 7850--7859 (2020)

\bibitem{tpd}
Yang, X., Ding, C., Hong, Z., Huang, J., Tao, J., Xu, X.: Texture-preserving
  diffusion models for high-fidelity virtual try-on. In: Proceedings of the
  IEEE/CVF Conference on Computer Vision and Pattern Recognition (CVPR). pp.
  7017--7026 (June 2024)

\bibitem{lpips}
Zhang, R., Isola, P., Efros, A.A., Shechtman, E., Wang, O.: The unreasonable
  effectiveness of deep features as a perceptual metric. In: Proceedings of the
  IEEE conference on computer vision and pattern recognition. pp. 586--595
  (2018)

\bibitem{boow}
Zhang, X., Song, D., Zhan, P., Chang, T., Zeng, J., Chen, Q., Luo, W., Liu,
  A.A.: Boow-vton: Boosting in-the-wild virtual try-on via mask-free pseudo
  data training. In: Proceedings of the Computer Vision and Pattern Recognition
  Conference. pp. 26399--26408 (2025)

\bibitem{charguid}
Zheng, C., Lan, Y.: Characteristic guidance: non-linear correction for
  diffusion model at large guidance scale. In: Proceedings of the 41st
  International Conference on Machine Learning. pp. 61386--61412 (2024)

\end{thebibliography}

\end{document}